\documentclass{article}

\usepackage{microtype}
\usepackage{graphicx}
\usepackage{booktabs}
\usepackage{multirow}
\usepackage{hyperref}

\usepackage[preprint]{icml2026}

\usepackage{amsmath,amssymb,amsthm}
\usepackage{url}
\usepackage{xcolor}

\theoremstyle{plain}
\newtheorem{theorem}{Theorem}
\newtheorem{proposition}[theorem]{Proposition}
\newtheorem{lemma}[theorem]{Lemma}
\theoremstyle{definition}
\newtheorem{definition}[theorem]{Definition}
\theoremstyle{remark}

\newcommand{\R}{\mathbb{R}}
\newcommand{\one}{\mathbf{1}}
\newcommand{\sm}{\sigma}
\newcommand{\Lip}{\operatorname{Lip}}
\newcommand{\simplex}{\Delta^{K-1}}
\newcommand{\elw}[1]{#1^{\odot}}

\icmltitlerunning{Order-Preserving Calibration with Input-Dependent Monotone Logit Maps}

\hypersetup{pdftitle={Order-Preserving Calibration with Input-Dependent Monotone Logit Maps},
  pdfauthor={Qipeng Zhan, Li Shen}, pdfsubject={}}

\begin{document}

\twocolumn[
  \icmltitle{Order-Preserving Calibration with\\
    Input-Dependent Monotone Logit Maps}

  \begin{icmlauthorlist}
    \icmlauthor{Qipeng Zhan}{upenn}
    \icmlauthor{Li Shen}{upenn}
  \end{icmlauthorlist}

  \icmlaffiliation{upenn}{University of Pennsylvania, Philadelphia, PA, USA}
  \icmlcorrespondingauthor{Qipeng Zhan}{qipengz@sas.upenn.edu}

  \icmlkeywords{Calibration, Uncertainty, Order Preservation, Monotone Networks}

  \vskip 0.3in
]

\printAffiliationsAndNotice{}

\begin{abstract}
Post-hoc calibrators that preserve the ranking of the classes on every input
can correct the confidence of a classifier without changing any of its
predictions. The most flexible of them apply a learned monotone function to
every logit, but
they have two shortcomings: monotonicity holds only approximately and, as we
show, can fail in practice; and the same function is applied to every input,
whether the classifier is confident or uncertain on it. Since order
preservation only constrains the logits within one input, each input can in
fact have its own map. We propose input-dependent monotone logit maps, which
add to a scaled identity a residual network that is contractive in the logit
and conditioned on a summary of the logit vector. The contraction makes
every map strictly increasing by construction, so accuracy is preserved
exactly. On CIFAR-10, CIFAR-100 and ImageNet, our method matches or improves
on existing post-hoc calibrators.
\end{abstract}

\section{Introduction}
\label{sec:intro}

A classifier is calibrated if its confidence agrees with its accuracy: among
the inputs to which it assigns probability $0.8$, about $80\%$ should be
classified correctly. Modern neural networks often fail this test, usually by
being overconfident \citep{guo2017calibration}, although how badly they fail
depends a great deal on the architecture and the training recipe
\citep{minderer2021revisiting}. Calibrated probabilities matter whenever a
prediction is weighed against costs, fed into another model, or used to
decide when to defer to a human. The cheapest fix is post-hoc calibration:
the classifier stays fixed, and a small map from its outputs, such as its
logits or its predicted probabilities, to new probabilities is fitted on
held-out data.

Temperature scaling \citep{guo2017calibration} divides all logits by a single
learned constant. It rarely overfits and never changes a prediction, but it
rescales every input, and every gap between two logits, by the same factor.
Vector and matrix scaling \citep{guo2017calibration}, Dirichlet calibration
\citep{kull2019dirichlet} and binning methods
\citep{zadrozny2001obtaining,zadrozny2002transforming} are more flexible, but
they can change the predicted class. In our experiments, histogram binning
raises the test error from $20.78\%$ to $22.22\%$ on CIFAR-100 and from
$17.74\%$ to $19.86\%$ on ImageNet.

If a calibrator keeps the ranking of the classes on every input, it cannot
change the accuracy \citep{rahimi2020intra}. The most expressive calibrators
of this kind are the diagonal calibrator (DIAG) of \citet{rahimi2020intra}
and the invertible logits transformation (InvLT) of \citet{zhao2026invlt}.
Both apply one scalar function to every logit, and both have two
limitations.

\emph{Monotonicity is not guaranteed.} DIAG integrates a positive derivative
network numerically \citep{wehenkel2019umnn}, so the function it computes is
monotone only up to quadrature error, and the integration is slow. InvLT
instead penalizes the reconstruction error of an inverse network trained
alongside it. This soft constraint keeps the function close to invertible on
the calibration data, but it does not rule out decreasing segments, and we
do observe them (Figure~\ref{fig:mapping}). In both methods, then, order
preservation, which is the reason for the construction in the first place,
holds only approximately.

\emph{The map is the same for every input.} Both methods fit a single
function for the whole dataset. A logit of a given value is therefore mapped
to the same output regardless of whether the classifier is confident on that
input or undecided between several classes, even though the rest of the
logit vector distinguishes the two cases. These cases can behave very
differently. Among CIFAR-100 test inputs whose top logit lies in
$[10, 11)$, the uncalibrated classifier is overconfident by $18$ points on
the half with the smaller gap to the second class, but only by $4$ points on
the other half (Figure~\ref{fig:gap-context}). Per-input temperature scaling
\citep{tomani2022pts,joy2023sample} does adapt to the input, but within a
given input it is still linear in the logits.

Our calibrator addresses both limitations. It preserves the order of the
logits exactly, and its map is nonlinear within an input and varies across
inputs. This is possible because order preservation only compares logits of
the same input, so each input can have its own monotone map as long as all
of its classes share it. For an input with logit vector $z$ we use
\begin{equation*}
  u \;\mapsto\; \alpha\,\big(u + h(u; v)\big), \qquad v = \kappa(z),
\end{equation*}
where $\alpha > 0$ is a fixed scale, $\kappa(z)$ is a permutation-invariant
summary of $z$ (for example its largest softmax probabilities), and $h$ is a
small network whose Lipschitz constant in $u$ is less than one. As in
invertible residual networks \citep{behrmann2019iresnet}, this bound makes
the map strictly increasing for every $v$; we call the result an
input-dependent monotone logit map. The bound is enforced by rescaling the
weights after each optimizer step. Monotonicity is therefore a hard
constraint that holds for every parameter value, and it needs neither
numerical integration nor an inverse network. The number of parameters does
not depend on the number of classes, and temperature scaling, logit-wise
monotone maps and per-input temperatures are all special cases. For large
label sets, we apply the map only to the $k$ largest logits and move the
remaining ones by a clipped common shift; this still preserves the order and
gives the exact likelihood.

We compare against twelve post-hoc calibrators on CIFAR-10, CIFAR-100 and
ImageNet. Our calibrator obtains the lowest negative log-likelihood and
expected calibration error on CIFAR-10 and CIFAR-100, lowering the ECE on
CIFAR-100 from $3.43\%$ for the best baseline to $1.89\%$, and it matches
the best baseline, InvLT, on ImageNet. We also find that the soft constraint
of InvLT is violated in practice. On ImageNet with $25000$ calibration
examples, four of its ten fitted maps decrease on an interval that contains
$32\%$ to $55\%$ of the test logits, and neither the accuracy nor the
calibration error gives any sign of it.

\section{Related work}
\label{sec:related}

\paragraph{Calibration.}
Calibration was first studied for probability forecasts
\citep{dawid1982well,degroot1983comparison}. \citet{guo2017calibration}
showed that deep classifiers are often miscalibrated, and
\citet{minderer2021revisiting} found that the extent varies across model
families. Remedies at training time include label smoothing
\citep{muller2019label}, focal loss \citep{mukhoti2020focal}, calibration
penalties \citep{kumar2018trainable}, mixup \citep{thulasidasan2019mixup},
ensembles \citep{lakshminarayanan2017simple} and Monte Carlo dropout
\citep{gal2016dropout}, all of which can be combined with post-hoc
calibration. Classical post-hoc methods are Platt scaling
\citep{platt1999probabilistic}, histogram binning
\citep{zadrozny2001obtaining}, isotonic regression
\citep{zadrozny2002transforming} and Bayesian binning
\citep{naeini2015obtaining} \citep[see also][]{niculescu2005predicting}.
Methods developed for neural networks include temperature, vector and matrix
scaling \citep{guo2017calibration}, Dirichlet calibration
\citep{kull2019dirichlet}, ensemble temperature scaling
\citep{zhang2020mixnmatch}, Gaussian-process calibration
\citep{wenger2020nonparametric}, mutual-information binning
\citep{patel2021multiclass}, splines \citep{gupta2021spline} and Meta-Cal
\citep{ma2021metacal}. Except for temperature scaling and its ensemble
variant, most of these can change the predicted class. Since binned
estimates of the expected calibration error are biased
\citep{kumar2019verified,nixon2019measuring,vaicenavicius2019evaluating},
we also report equal-mass, classwise and kernel estimates.

\paragraph{Order-preserving and input-dependent calibrators.}
\citet{rahimi2020intra} introduced intra order-preserving functions. Their
diagonal subfamily applies one monotone scalar function, built from
unconstrained monotonic networks \citep{wehenkel2019umnn}, to every logit,
and they also propose an order-invariant subfamily. Their general
construction takes the sorted logit vector as input, so its size grows with
the number of classes. InvLT \citep{zhao2026invlt} learns one scalar
function together with an inverse network. Neither DIAG nor InvLT guarantees
monotonicity (DIAG has it up to quadrature error, InvLT encourages it with a
penalty), and both apply the same function to the logits of every input.
Other methods predict a temperature for each input, either from the logits
\citep{tomani2022pts} or from learned features \citep{joy2023sample}, or for
each pixel in segmentation \citep{ding2021local}. MCCT \citep{zhang2025mcct}
sorts the logits and applies an affine map whose scale and shift depend on
the rank and are constrained so that the order is kept. Our calibrator
includes DIAG-type and
per-input temperature calibrators as special cases, but unlike them, its
scalar function is nonlinear within an input and changes from one input to
the next.

\paragraph{Monotone and Lipschitz networks.}
Monotone networks have been built from min--max compositions of linear units
\citep{sill1998monotonic}, from lattices \citep{you2017deep}, by integrating
positive networks \citep{wehenkel2019umnn}, with certificates computed after
training \citep{liu2020certified}, and with constrained weights and
activations \citep{runje2023constrained}. \citet{nolte2023expressive} add a
linear term to a network with bounded Lipschitz constant; the sum is
monotone as long as the linear term dominates. In one dimension, the same
idea underlies the invertibility of residual networks
\citep{behrmann2019iresnet,chen2019residual}: $u \mapsto u + h(u)$ is a
bijection when $\Lip(h) < 1$. Lipschitz bounds are commonly enforced with
spectral normalization \citep{miyato2018spectral}, by projecting the weights
onto norm balls \citep{gouk2021regularisation}, or with norm-preserving
activations \citep{anil2019sorting}. We bound the Lipschitz constant only
with respect to the logit and leave the dependence on the context
unconstrained.

\section{Preliminaries}
\label{sec:setup}

Let $K \geq 2$ be the number of classes, $z = (z_1,\dots,z_K)^\top \in \R^K$
the logit vector that a trained classifier produces on an input, and
$\one = (1,\dots,1)^\top \in \R^K$. The softmax $\sm : \R^K \to \R^K$ is
\begin{equation}
  \sm(z)_i \;=\; \frac{e^{z_i}}{\sum_{j=1}^K e^{z_j}},
  \qquad i = 1,\dots,K,
\end{equation}
and $\simplex = \{p \in \R^K : p_i \geq 0,\ \sum_i p_i = 1\}$ is the
probability simplex. For a scalar function $g : \R \to \R$, we write
$\elw{g} : \R^K \to \R^K$ for its coordinate-wise extension,
\begin{equation}
  \label{eq:elw}
  \elw{g}(z) \;:=\; \big(g(z_1),\, g(z_2),\, \dots,\, g(z_K)\big)^\top ,
\end{equation}
and keep the plain symbol $g$ for the scalar function.

A \emph{post-hoc calibrator} replaces the predicted distribution $\sm(z)$ of
a fixed classifier by a new distribution in $\simplex$. It is fitted on a
held-out set and may act on different outputs of the classifier: binning
methods and isotonic regression, for example, transform the predicted
probabilities, whereas temperature scaling and Dirichlet calibration
transform the logits. The goal is that, among the inputs on which the
calibrated model gives probability $p$ to its predicted class, a fraction
close to $p$ is classified correctly.

In this paper we consider calibrators that act on the logits, that is, maps
$\Phi : \R^K \to \R^K$ with calibrated prediction $\sm(\Phi(z))$, usually
fitted by minimizing the negative log-likelihood of $\sm(\Phi(z))$. In
addition, we want the calibrator to keep the ranking of the classes on every
input. Since the softmax preserves the order of its arguments, we can state
this condition in terms of the logits.

\begin{definition}[Order preservation]
\label{def:order}
A calibrator $\Phi : \R^K \to \R^K$ is \emph{order preserving} if
\begin{equation}
  \label{eq:order}
  \begin{aligned}
  &\forall z \in \R^K,\ \forall i,j \in \{1,\dots,K\} : \\
  &\qquad z_i \geq z_j \iff \Phi(z)_i \geq \Phi(z)_j .
  \end{aligned}
\end{equation}
\end{definition}

Condition \eqref{eq:order} preserves the preorder that the logits induce on
the classes. In particular, $\arg\max_i \Phi(z)_i = \arg\max_i z_i$ for every
$z$, so the calibrated model makes the same predictions as the base
classifier and has the same accuracy. Such a calibrator can therefore be
added to a deployed model without re-evaluating its accuracy. Order
preservation is stronger than keeping the predicted class alone: it also
keeps every top-$k$ prediction and any statistic that depends only on the
ranking of the classes.

\paragraph{Logit-wise calibrators.}
A simple way to get an order-preserving calibrator that is more flexible than
temperature scaling is to apply one scalar function $\phi : \R \to \R$ to
every logit,
\begin{equation}
  \label{eq:logitwise}
  \begin{aligned}
  \Phi_\phi &\;=\; \elw{\phi}, \qquad\text{that is,} \\
  \Phi_\phi(z) &\;=\; \big(\phi(z_1),\, \phi(z_2),\, \dots,\, \phi(z_K)\big)^\top .
  \end{aligned}
\end{equation}
We call such calibrators \emph{logit-wise}. Since $\phi$ is shared by all
classes, the number of parameters does not depend on $K$, and $\Phi_\phi$ is
order preserving if and only if $\phi$ is strictly increasing. Temperature
scaling \citep{guo2017calibration} corresponds to $\phi(t) = at$ with inverse
temperature $a > 0$. DIAG \citep{rahimi2020intra} computes $\phi$ by
numerically integrating a positive network \citep{wehenkel2019umnn}. Each
value $\phi(z_i)$ thus comes from a separate quadrature, and two close
logits are guaranteed to be ordered correctly only if the quadrature error
is smaller than their gap. InvLT \citep{zhao2026invlt} adds to the loss the
reconstruction error of an inverse network trained jointly with $\phi$; this
encourages $\phi$ to be monotone but does not enforce it.

\begin{figure}[t]
\centering
\includegraphics[width=\linewidth]{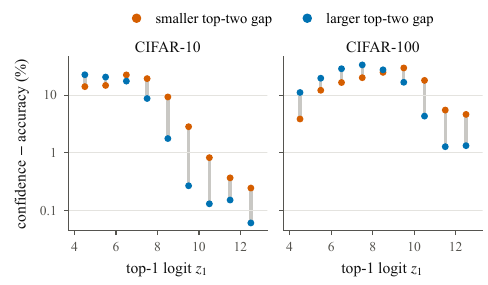}
\caption{Overconfidence of the uncalibrated classifier among test inputs with
the same top logit, pooled over the $10$ seeds. Inputs are grouped by $z_1$
in bins of width one, groups with fewer than $1000$ inputs are omitted, and
each group is split at the median gap between the two largest softmax
probabilities. Each pair of points shows confidence minus accuracy for the
two halves, on a log scale.}
\label{fig:gap-context}
\end{figure}

\paragraph{One map per input.}
Condition \eqref{eq:order} relates the coordinates of $\Phi(z)$ to those of
the same $z$; it says nothing about how the outputs on two different inputs
should relate. A logit-wise calibrator nevertheless uses the same $\phi$
everywhere, even though the rest of the logit vector can tell a confident
input from an undecided one. Figure~\ref{fig:gap-context} shows why this
matters for the uncalibrated ResNet-50 classifiers of
Section~\ref{sec:experiments}. We group the test inputs by their top logit
$z_1$ and split each group at the median gap between the two largest softmax
probabilities. Inputs with the same top logit can differ widely in
overconfidence. For $z_1 \in [9, 10)$ on CIFAR-10, the half with the smaller
gap is overconfident by $2.8$ points and the other half by $0.3$; for
$z_1 \in [10, 11)$ on CIFAR-100, the numbers are $18.1$ and $4.3$. A
logit-wise calibrator has to apply the same function to both halves. Our
calibrator still uses a single strictly increasing map per input, shared by
all of its classes, but lets each input choose that map.

\section{Method}
\label{sec:method}

\subsection{Monotone maps from contractive residuals}
\label{sec:residual}

A logit-wise calibrator is order preserving exactly when its scalar map is
strictly increasing. Rather than constraining $\phi$ directly, we write it as
the identity plus a residual, $\phi = \mathrm{id}_\R + h$, and constrain the
residual.

\begin{lemma}[Residual monotone maps]
\label{lem:residual}
Let $h : \R \to \R$ be Lipschitz with $\Lip(h) \leq L < 1$, and let
$\phi(u) = u + h(u)$. Then for all $u \geq u'$
\begin{equation}
  \label{eq:bilip}
  (1 - L)\,(u - u') \;\leq\; \phi(u) - \phi(u') \;\leq\; (1 + L)\,(u - u') .
\end{equation}
In particular $\phi$ is strictly increasing and a bijection of $\R$, and its
inverse is $(1-L)^{-1}$-Lipschitz.
\end{lemma}

The proof is in Appendix~\ref{app:proofs}.

Lemma~\ref{lem:residual} is the scalar case of the invertibility result for
residual networks \citep{behrmann2019iresnet}. It reduces monotonicity to a
single condition, $\Lip(h) < 1$, which we impose exactly on the weights of
$h$ (Section~\ref{sec:conditioned}). This is a hard constraint, unlike the
quadrature in DIAG or the penalty in InvLT. Monotonicity, and with it
accuracy preservation, holds for every parameter value, at every step of
training and on every input, including logits outside the range seen in the
calibration data.

By \eqref{eq:bilip}, the slopes of $\phi$ lie in $[1-L, 1+L]$. Multiplying by
a constant $\alpha > 0$ keeps the map strictly increasing and rescales this
range:
\begin{equation}
  \label{eq:bilipscaled}
  \begin{aligned}
  \alpha(1 - L)\,(u - u') &\;\leq\; \alpha\phi(u) - \alpha\phi(u') \\
  &\;\leq\; \alpha(1 + L)\,(u - u') \qquad (u \geq u').
  \end{aligned}
\end{equation}
The scale $\alpha$ sets the average slope, while the residual $h$ only
controls how far the map departs from a straight line. The ratio between any
two slopes of $\alpha\phi$ is at most $(1+L)/(1-L)$.

\subsection{Input-dependent monotone logit maps}
\label{sec:conditioned}

Let $\kappa : \R^K \to \R^m$ be a context map that summarizes the logit
vector by $m$ features, with $m$ fixed. A learned residual
\begin{equation*}
  h : \R \times \R^m \to \R, \qquad (u; v) \mapsto h(u; v)
\end{equation*}
takes the logit of one class as its first argument and the context
$v = \kappa(z)$ of the whole input as its second. The scale $\alpha > 0$ is a
hyperparameter shared by all inputs. The calibrator is
\begin{equation}
  \label{eq:method}
  \boxed{\;
  \Phi(z)_i \;=\; \alpha\,
  \Big( z_i + h\big(z_i;\, \kappa(z)\big) \Big),
  \qquad i = 1,\dots,K, \;}
\end{equation}
and the calibrated distribution is $\sm(\Phi(z))$. The only constraint we
place on $h$ is
\begin{equation}
  \label{eq:lipconstraint}
  \sup_{v \in \R^m} \Lip\big(h(\,\cdot\,; v)\big) \;\leq\; L \;<\; 1,
\end{equation}
that is, $u \mapsto h(u; v)$ must be $L$-Lipschitz for every fixed context
$v$. How $h$ depends on $v$ is left unconstrained.

\begin{figure}[t]
\centering
\colorlet{cA}{blue!80!black}
\colorlet{cB}{orange!90!black}
\colorlet{cC}{green!50!black}
\colorlet{cD}{red!75!black}
\setlength{\unitlength}{1cm}%
\resizebox{\columnwidth}{!}{%
\begin{picture}(11.2,1.9)(-2.6,-0.35)%
  \color{gray!35}%
  \put(-1.8,1.2){\line(1,0){7}}%
  \put(-1.8,0){\line(1,0){7}}%
  {\color{cA!60}\qbezier(-1.45,1.2)(-1.225,0.6)(-1,0)}%
  {\color{cB!60}\qbezier(-0.45,1.2)(-0.325,0.6)(-0.2,0)}%
  {\color{cC!60}\qbezier(0.15,1.2)(0.2,0.6)(0.25,0)}%
  {\color{cD!60}\qbezier(1.75,1.2)(1.425,0.6)(1.1,0)}%
  {\color{cA}\put(-1.45,1.2){\circle*{0.16}}\put(-1,0){\circle*{0.16}}}%
  {\color{cB}\put(-0.45,1.2){\circle*{0.16}}\put(-0.2,0){\circle*{0.16}}}%
  {\color{cC}\put(0.15,1.2){\circle*{0.16}}\put(0.25,0){\circle*{0.16}}}%
  {\color{cD}\put(1.75,1.2){\circle*{0.16}}\put(1.1,0){\circle*{0.16}}}%
  \color{black}%
  \put(5.35,1.2){\makebox(0,0)[l]{$z$}}%
  \put(5.35,0){\makebox(0,0)[l]{$\Phi(z) = \elw{(\phi_v)}(z)$}}%
  \put(-1.95,0.6){\makebox(0,0)[r]{$\phi_v$}}%
\end{picture}}
\caption{The calibrator \eqref{eq:method} on one input with $K = 4$ classes.
The map $\phi_v$, selected by the context $v = \kappa(z)$, is strictly
increasing, so the segments do not cross: the gaps change but the ranking
does not. A different input would get a different map.}
\label{fig:method}
\end{figure}
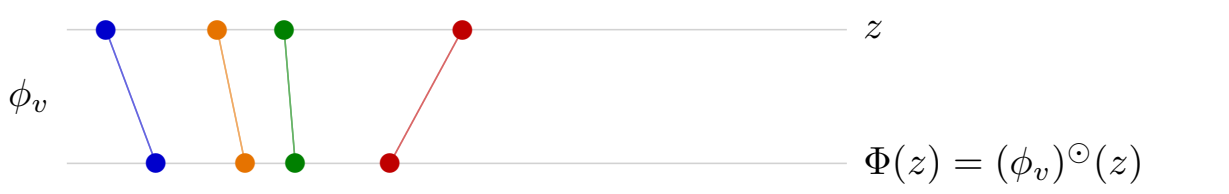

Let $\phi_v(u) := \alpha\,\big(u + h(u; v)\big)$. Then \eqref{eq:method} can
be written as $\Phi(z) = \elw{(\phi_v)}(z)$ with $v = \kappa(z)$. Within one
input, a single strictly increasing scalar map is applied to all $K$ logits,
just as in a logit-wise calibrator; the difference is that the input selects
this map through its context. We call $\Phi$ an \emph{input-dependent
monotone logit map}.
Figure~\ref{fig:method} illustrates the construction.

\paragraph{Order preservation.}
The guarantee holds for any context map, any positive scale and any residual
that satisfies \eqref{eq:lipconstraint}.

\begin{proposition}[Order preservation]
\label{prop:order}
If $h$ satisfies \eqref{eq:lipconstraint} and $\alpha > 0$, then the
calibrator $\Phi$ of \eqref{eq:method} is order preserving for every
$\kappa : \R^K \to \R^m$. In particular
$\arg\max_i \Phi(z)_i = \arg\max_i z_i$ for every $z$, and the accuracy of the
base classifier is unchanged.
\end{proposition}

\begin{proof}
Fix $z$ and let $v = \kappa(z)$. By \eqref{eq:lipconstraint} and
Lemma~\ref{lem:residual}, $u \mapsto u + h(u; v)$ is strictly increasing, and
so is $\phi_v$ because $\alpha > 0$. Since $\phi_v$ is applied to every
coordinate of $z$,
$\Phi(z)_i \geq \Phi(z)_j \iff \phi_v(z_i) \geq \phi_v(z_j) \iff
z_i \geq z_j$.
\end{proof}

The proof uses the context only through the fact that it is shared by all
$K$ coordinates of one input; it makes no assumption about $\kappa$ or about
how $h$ depends on $v$. We can therefore choose the context purely for
calibration quality, with no risk to accuracy.

\paragraph{Choice of context and network.}
We take $\kappa(z)$ to be a sorted top-$m$ summary of the input, either the
$m$ largest softmax probabilities or the $m$ largest logits, in decreasing
order, with $m \leq K$. By default we use probabilities, since they are
bounded in $[0, 1]$, do not change when all logits are shifted by the same
constant, and work best on ImageNet, our largest dataset. Logits work better
on CIFAR (Appendix~\ref{app:context}). With our default $m = 2$, the context
contains the gap between the two largest probabilities, which is the
quantity used to split each group in Figure~\ref{fig:gap-context}. For both
kinds of context, the width $m$ does not depend on $K$, and hence neither
does the number of parameters of $h$. Both are also invariant to
permutations of the classes, so $h$ never sees class identities and $\Phi$
commutes with any relabeling: $\Phi(Pz) = P\,\Phi(z)$ for every permutation
matrix $P$. The residual $h$ is a multilayer perceptron applied to the
concatenation $(u, v) \in \R^{1+m}$. We initialize its output layer to zero,
so training starts from $h \equiv 0$, i.e., from temperature scaling with
fixed temperature $1/\alpha$, $\Phi(z) = \alpha z$.

\paragraph{Enforcing the Lipschitz constraint.}
Constraint \eqref{eq:lipconstraint} involves only the first argument of $h$,
and it can be enforced exactly. Write the first layer of the perceptron as
$w\,u + V v + b_1$ with $w \in \R^{n_1}$, let $W_2, \dots, W_\ell$ be the
remaining weight matrices, and use a $1$-Lipschitz activation such as the
ReLU. For fixed $v$, changing $u$ to $u'$ changes the first-layer
pre-activation by $w\,(u - u')$, and each subsequent layer amplifies a
change by at most the spectral norm of its weight matrix. Hence
\begin{equation*}
  \Lip\big(h(\,\cdot\,; v)\big) \;\leq\; \|w\|_2 \prod_{l=2}^{\ell} \|W_l\|_2
  \qquad \text{for every } v \in \R^m .
\end{equation*}
It is enough to keep this product at most $L$. After every optimization step
we rescale $w$ so that $\|w\|_2 \leq 1$, each hidden matrix so that
$\|W_l\|_2 \leq 1$, and the output layer so that $\|W_\ell\|_2 \leq L$, in
each case only if the norm exceeds the bound. This is similar to spectral
normalization \citep{miyato2018spectral}, but because the matrices are
small, we compute their spectral norms exactly instead of by power
iteration. The context weights $V$ are not constrained.

\paragraph{Role of the scale and the residual.}
Applying \eqref{eq:bilipscaled} to $\phi_v$ gives, whenever $z_i \geq z_j$,
\begin{equation*}
  \alpha(1 - L)\,(z_i - z_j) \;\leq\; \Phi(z)_i - \Phi(z)_j \;\leq\;
  \alpha(1 + L)\,(z_i - z_j).
\end{equation*}
With $\alpha = 1$, a gap could shrink to a fraction $1-L$ of its size but
could not even double, which leaves little room to sharpen an underconfident
classifier. The scale $\alpha$ sets the overall stretch, and the residual
varies it within $[\alpha(1-L), \alpha(1+L)]$, both across the gaps of one
input and across inputs. In other words, the calibrator is a global
temperature $1/\alpha$ combined with a bounded, input-dependent nonlinear
correction.

\paragraph{Special cases.}
Several known calibrators are obtained from particular residuals that
satisfy \eqref{eq:lipconstraint}.
\begin{itemize}
  \setlength{\itemsep}{0pt}
  \item A residual that ignores the context, $h(u; v) = h(u)$, gives the
        logit-wise calibrator $\Phi = \elw{\phi}$ of \eqref{eq:logitwise} with
        $\phi = \alpha\,(\mathrm{id}_\R + h)$. Its monotonicity follows from
        Lemma~\ref{lem:residual}, and its difference quotients lie in
        $[\alpha(1-L), \alpha(1+L)]$.
  \item $h \equiv 0$ gives $\Phi(z) = \alpha z$, temperature scaling with
        temperature $1/\alpha$. Training starts from this point.
  \item $h(u; v) = (a(v) - 1)\,u$ with a learned
        $a : \R^m \to [1-L, 1+L]$ gives the input-dependent temperature
        $1/(\alpha\, a(v))$, as in parameterized temperature scaling
        \citep{tomani2022pts}. Here $\alpha$ sets the level of the
        temperature and $a(v)$ varies it by a factor in $[1-L, 1+L]$.
\end{itemize}
In general, however, the map is nonlinear within an input, so the gaps among
the leading classes can be rescaled differently from those among the
trailing ones, and the map changes with the context from one input to the
next.

\begin{figure*}[t]
\centering
\includegraphics[width=\textwidth]{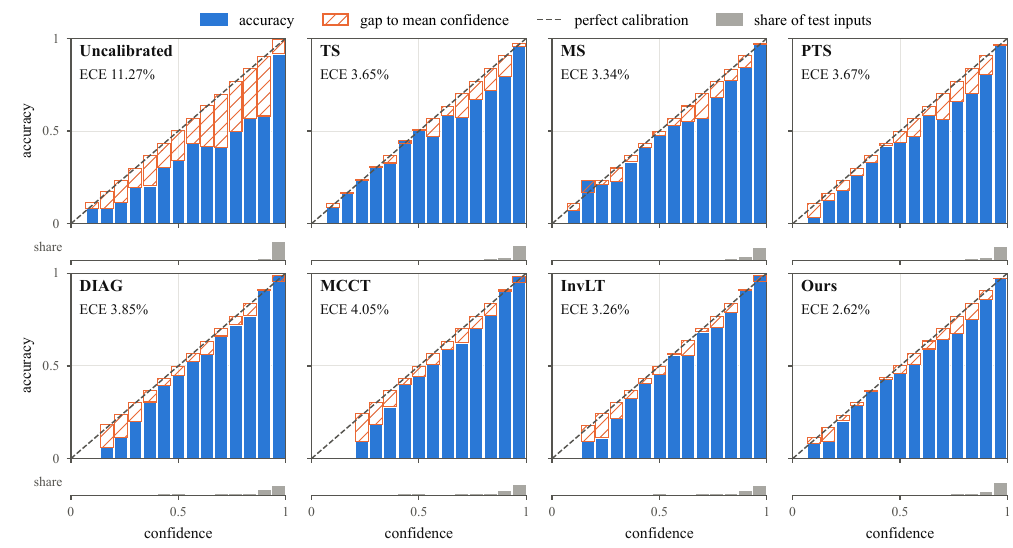}
\caption{Reliability diagrams on CIFAR-100 with $5000$ calibration examples
and seed $0$, with $15$ equal-width bins. Bins with fewer than $10$ test
inputs are not drawn; the ECE in each panel uses all $15$ bins.}
\label{fig:reliability-main}
\end{figure*}

\subsection{Many classes: calibrating only the top-\texorpdfstring{$k$}{k} logits}
\label{sec:topk}

The calibrator \eqref{eq:method} evaluates the residual network once per
class. When $K$ is large, as on ImageNet-1k \citep{deng2009imagenet} with
$K = 1000$, most of this computation goes to classes with negligible
probability, whereas the confidence of the prediction and the competition
among the leading classes depend on only a few of the largest logits. We
therefore apply the input-dependent map only to the $k$ largest logits,
which we call the \emph{head}, and move the remaining logits, the
\emph{tail}, by a common affine map.

\paragraph{Construction.}
Let $\tau(z)$ be the $k$-th largest entry of $z$, $v = \kappa(z)$, and
$c(z) := h(\tau(z); v)$ the residual at the $k$-th logit. The top-$k$
calibrator is
\begin{equation}
  \label{eq:topk}
  \Phi^{k}(z)_i \;=\;
  \begin{cases}
    \alpha\,\big(z_i + h(z_i; v)\big), & z_i \geq \tau(z), \\[2pt]
    \alpha\,\big(z_i + \min\{c(z), 0\}\big), & z_i < \tau(z).
  \end{cases}
\end{equation}
The head is calibrated as in \eqref{eq:method}. The tail is scaled by
$\alpha$ and, if the residual at the $k$-th logit is negative, shifted by
that residual; otherwise it is not shifted. For $k = K$ the tail is empty
and $\Phi^{K} = \Phi$. During training we stop the gradient through the
tail's copy of $c(z)$, so the network is fitted through the head only.
Without this, the gradients of all $K - k$ tail coordinates pile up on a
single output of the network, which made training unstable in our
experiments.

\begin{proposition}[Order preservation of the top-$k$ calibrator]
\label{prop:topk}
Under the assumptions of Proposition~\ref{prop:order}, $\Phi^{k}$ is order
preserving for every $k \in \{1, \dots, K\}$ and every context map $\kappa$.
\end{proposition}

\noindent The proof, in Appendix~\ref{app:proofs}, compares a head logit with
a tail logit through the residual at the $k$-th logit.

\paragraph{Cost and exact likelihood.}
The residual network is evaluated at $k$ logits instead of $K$, ten times
fewer on ImageNet-1k with $k = 100$. The likelihood is still computed
exactly over all $K$ classes. The tail enters the normalizing constant of
the softmax only through
$e^{\alpha \min\{c(z), 0\}} \sum_{j:\, z_j < \tau(z)} e^{\alpha z_j}$, and
because $\alpha$ is fixed, the sum is a per-example constant that can be
precomputed once before training. Renormalizing the softmax over the top $k$
classes would be even cheaper, but it would fit a different distribution
from the one that is deployed, and it is undefined when the label falls in
the tail. Appendix~\ref{app:topk} measures the speed and accuracy of this
approximation on ImageNet.

\paragraph{Clipping the tail shift.}
The proof of Proposition~\ref{prop:topk} requires the common tail shift $s$
to satisfy $s \leq c(z)$; otherwise a tail logit just below $\tau(z)$ could
overtake the $k$-th calibrated logit. The unclipped choice $s = c(z)$ meets
this requirement, but when $c(z) > 0$ it raises the tail together with the
head, and in our experiments the tail then got too much probability. The
problem is that $c(z)$ is fitted to reshape the head near its smallest
logit, yet it multiplies the mass of all $K - k$ tail classes by
$e^{\alpha c(z)}$ at once. With $s = \min\{c(z), 0\}$, the tail is moved
down when the order requires it and otherwise stays where temperature
scaling would put it. Consequently, the probability of each tail class
relative to the $k$-th class is never larger than under temperature scaling
with temperature $1/\alpha$. Among all shifts with $s \leq c(z)$, this is
also the one closest to $s = 0$.

\section{Experiments}
\label{sec:experiments}

\begin{table*}[t]
\centering
\caption{Calibration results on CIFAR with $5000$ calibration examples, mean
and standard deviation over $10$ seeds. All ECE variants are in \%. Best mean
among the calibrators in \textbf{bold}, second best \underline{underlined};
the uncalibrated model is excluded from the ranking. ImageNet results are in
Table~\ref{tab:imagenet}.}
\label{tab:main}
\small
\resizebox{\textwidth}{!}{%
\begin{tabular}{l ccccc ccccc}
\toprule
& \multicolumn{5}{c}{CIFAR-10 (ResNet-50)}
& \multicolumn{5}{c}{CIFAR-100 (ResNet-50)} \\
\cmidrule(lr){2-6}\cmidrule(lr){7-11}
Method & NLL & ECE & AdaECE & CECE & KDE-ECE
       & NLL & ECE & AdaECE & CECE & KDE-ECE \\
\midrule
Uncalibrated & 0.209$_{\pm 0.007}$ & 3.13$_{\pm 0.12}$ & 3.12$_{\pm 0.12}$ & 0.663$_{\pm 0.024}$ & 4.29$_{\pm 0.18}$ & 0.976$_{\pm 0.014}$ & 11.77$_{\pm 0.42}$ & 11.76$_{\pm 0.42}$ & 0.266$_{\pm 0.007}$ & 14.13$_{\pm 0.52}$ \\
\midrule
TS & 0.166$_{\pm 0.005}$ & 0.96$_{\pm 0.22}$ & 1.27$_{\pm 0.13}$ & 0.348$_{\pm 0.036}$ & 1.47$_{\pm 0.25}$ & 0.876$_{\pm 0.010}$ & 4.44$_{\pm 0.39}$ & 4.34$_{\pm 0.39}$ & 0.172$_{\pm 0.004}$ & 4.67$_{\pm 0.41}$ \\
VS & 0.161$_{\pm 0.004}$ & 0.94$_{\pm 0.25}$ & 1.08$_{\pm 0.20}$ & 0.289$_{\pm 0.032}$ & 1.36$_{\pm 0.29}$ & 0.880$_{\pm 0.012}$ & 4.71$_{\pm 0.39}$ & 4.59$_{\pm 0.39}$ & 0.163$_{\pm 0.007}$ & 5.09$_{\pm 0.41}$ \\
MS & 0.158$_{\pm 0.004}$ & 0.91$_{\pm 0.26}$ & 1.07$_{\pm 0.21}$ & \textbf{0.288}$_{\pm 0.030}$ & 1.37$_{\pm 0.28}$ & 0.825$_{\pm 0.010}$ & 3.81$_{\pm 0.38}$ & 3.70$_{\pm 0.42}$ & \textbf{0.153}$_{\pm 0.005}$ & 4.07$_{\pm 0.40}$ \\
HB & 0.350$_{\pm 0.041}$ & 0.86$_{\pm 0.21}$ & 2.28$_{\pm 0.19}$ & 0.359$_{\pm 0.027}$ & \underline{0.98}$_{\pm 0.18}$ & 2.179$_{\pm 0.078}$ & 6.53$_{\pm 0.52}$ & 8.49$_{\pm 0.48}$ & 0.179$_{\pm 0.007}$ & 7.01$_{\pm 0.33}$ \\
IR & 0.214$_{\pm 0.012}$ & 1.07$_{\pm 0.25}$ & 0.86$_{\pm 0.31}$ & 0.319$_{\pm 0.040}$ & 1.50$_{\pm 0.38}$ & 1.300$_{\pm 0.061}$ & 6.17$_{\pm 0.38}$ & 6.09$_{\pm 0.36}$ & 0.178$_{\pm 0.007}$ & 7.07$_{\pm 0.41}$ \\
Spline & 0.245$_{\pm 0.022}$ & 1.69$_{\pm 0.19}$ & 1.72$_{\pm 0.13}$ & 0.409$_{\pm 0.032}$ & 2.30$_{\pm 0.27}$ & 0.987$_{\pm 0.017}$ & 7.66$_{\pm 0.28}$ & 7.70$_{\pm 0.27}$ & 0.193$_{\pm 0.006}$ & 8.14$_{\pm 0.29}$ \\
Dirichlet & 0.160$_{\pm 0.004}$ & 0.93$_{\pm 0.27}$ & 1.14$_{\pm 0.21}$ & \textbf{0.288}$_{\pm 0.033}$ & 1.42$_{\pm 0.27}$ & 0.830$_{\pm 0.011}$ & 3.83$_{\pm 0.40}$ & 3.70$_{\pm 0.47}$ & 0.157$_{\pm 0.006}$ & 4.09$_{\pm 0.42}$ \\
ETS & 0.166$_{\pm 0.005}$ & 0.96$_{\pm 0.22}$ & 1.27$_{\pm 0.13}$ & 0.348$_{\pm 0.036}$ & 1.47$_{\pm 0.25}$ & 0.876$_{\pm 0.010}$ & 4.44$_{\pm 0.39}$ & 4.34$_{\pm 0.39}$ & 0.172$_{\pm 0.004}$ & 4.67$_{\pm 0.41}$ \\
PTS & 0.163$_{\pm 0.005}$ & 0.87$_{\pm 0.25}$ & 0.84$_{\pm 0.23}$ & 0.339$_{\pm 0.032}$ & 1.30$_{\pm 0.30}$ & 0.872$_{\pm 0.010}$ & 4.32$_{\pm 0.35}$ & 4.55$_{\pm 0.42}$ & 0.172$_{\pm 0.004}$ & 4.58$_{\pm 0.36}$ \\
DIAG & 0.158$_{\pm 0.004}$ & 0.75$_{\pm 0.24}$ & 1.02$_{\pm 0.28}$ & 0.321$_{\pm 0.037}$ & 1.20$_{\pm 0.23}$ & 0.792$_{\pm 0.009}$ & 3.76$_{\pm 0.41}$ & 4.15$_{\pm 0.39}$ & 0.162$_{\pm 0.002}$ & 3.91$_{\pm 0.33}$ \\
MCCT & 0.220$_{\pm 0.073}$ & 0.79$_{\pm 0.24}$ & 1.29$_{\pm 0.14}$ & 0.329$_{\pm 0.036}$ & 1.39$_{\pm 0.18}$ & 0.981$_{\pm 0.236}$ & 3.84$_{\pm 0.54}$ & 4.20$_{\pm 0.40}$ & 0.165$_{\pm 0.004}$ & 3.91$_{\pm 0.40}$ \\
InvLT & \underline{0.156}$_{\pm 0.004}$ & \underline{0.69}$_{\pm 0.19}$ & \underline{0.82}$_{\pm 0.22}$ & 0.308$_{\pm 0.028}$ & 1.10$_{\pm 0.24}$ & \underline{0.790}$_{\pm 0.007}$ & \underline{3.43}$_{\pm 0.36}$ & \underline{3.66}$_{\pm 0.29}$ & 0.160$_{\pm 0.004}$ & \underline{3.48}$_{\pm 0.31}$ \\
\midrule
Ours & \textbf{0.155}$_{\pm 0.004}$ & \textbf{0.60}$_{\pm 0.23}$ & \textbf{0.54}$_{\pm 0.24}$ & 0.301$_{\pm 0.032}$ & \textbf{0.92}$_{\pm 0.30}$ & \textbf{0.778}$_{\pm 0.007}$ & \textbf{1.89}$_{\pm 0.49}$ & \textbf{2.15}$_{\pm 0.61}$ & \textbf{0.153}$_{\pm 0.003}$ & \textbf{2.22}$_{\pm 0.59}$ \\
\bottomrule
\end{tabular}}
\end{table*}

\subsection{Setup}
\label{sec:exp-setup}

\paragraph{Data and metrics.}
We calibrate ResNet-50 models \citep{he2016resnet} trained on CIFAR-10 and
CIFAR-100 \citep{krizhevsky2009cifar}, and a pretrained ResNet-152 on
ImageNet-1k \citep{deng2009imagenet}. Each calibrator is fitted on $5000$
held-out examples for CIFAR and $25000$ for ImageNet, with the classifier
kept fixed, and then evaluated on the test set. We report the negative
log-likelihood (NLL), the expected calibration error with
$10$ equal-width bins (ECE) \citep{guo2017calibration}, its equal-mass variant
(AdaECE) \citep{nguyen2015posterior,mukhoti2020focal}, the classwise ECE
(CECE) \citep{kull2019dirichlet}, and a binning-free kernel estimate of the
ECE with a Gaussian kernel of bandwidth $0.02$ (KDE-ECE)
\citep{zhang2020mixnmatch}; Appendix~\ref{app:metrics} gives the
definitions. All numbers are averaged over $10$ seeds.

\paragraph{Baselines.}
We compare with temperature, vector and matrix scaling (TS, VS, MS)
\citep{guo2017calibration}, histogram binning (HB)
\citep{zadrozny2001obtaining}, isotonic regression (IR)
\citep{zadrozny2002transforming}, spline calibration
\citep{gupta2021spline}, Dirichlet calibration \citep{kull2019dirichlet},
ensemble temperature scaling (ETS) \citep{zhang2020mixnmatch},
parameterized temperature scaling (PTS) \citep{tomani2022pts}, the diagonal
intra order-preserving calibrator with UMNN (DIAG) \citep{rahimi2020intra},
MCCT \citep{zhang2025mcct} and InvLT \citep{zhao2026invlt}.

\paragraph{Implementation.}
The residual $h$ is a perceptron with two hidden layers of width $10$ and
ReLU activations. To keep the setup the same across datasets, the context is
always the vector of the $2$ largest softmax probabilities, in decreasing
order. With the $2$ largest logits instead, the CIFAR-100 ECE would be
$1.23\%$ rather than $1.89\%$ (Appendix~\ref{app:context}). We use
$L = 0.99$, enforce the Lipschitz bound after every step as described in
Section~\ref{sec:conditioned}, and fix $\alpha = 2$ on CIFAR and $\alpha = 4$
on ImageNet. In Tables~\ref{tab:main} and~\ref{tab:imagenet} the map is
applied to all logits; Section~\ref{sec:exp-topk} evaluates the top-$k$
calibrator of Section~\ref{sec:topk} with $k = 100$ on ImageNet. We train
with Adam for $100$ epochs, using mini-batches of $5000$ examples (the whole
calibration set on CIFAR and a fifth of it on ImageNet), learning rate $0.1$
and cosine decay.

\subsection{Main results}
\label{sec:exp-main}

Table~\ref{tab:main} summarizes the results on CIFAR, and
Table~\ref{tab:imagenet} in Appendix~\ref{app:imagenet} those on ImageNet.
Like the other order-preserving methods (TS, ETS, PTS, DIAG, MCCT, InvLT),
our calibrator leaves the test error of the base classifier unchanged on all
three datasets. The remaining calibrators can change it, histogram binning
most of all: it raises the error from $20.78\%$ to $22.22\%$ on CIFAR-100
and from $17.74\%$ to $19.86\%$ on ImageNet.

On CIFAR-10 and CIFAR-100, our calibrator has the best NLL, ECE, AdaECE and
KDE-ECE. On CIFAR-100 it also has the best CECE, tied with MS, while on
CIFAR-10 the class-wise methods VS, MS and Dirichlet do slightly better on
this metric. The largest improvement is on CIFAR-100, where the ECE drops
from $3.43\%$ for the best baseline to $1.89\%$, with similar reductions in
AdaECE and KDE-ECE. Figure~\ref{fig:reliability-main} shows reliability
diagrams on CIFAR-100 for one seed. Temperature scaling, matrix scaling and
PTS remain overconfident for confidences between about $0.5$ and $0.9$.
DIAG, MCCT and InvLT reduce this gap but are overconfident on the least
confident inputs. Our calibrator has small gaps on the least confident
inputs and a smaller gap than temperature scaling in the middle range; in
this run its ECE is $2.62\%$, compared with $3.26\%$ for InvLT.

On ImageNet, the three monotone logit maps perform similarly. Our calibrator
has the lowest ECE ($0.39\%$, compared with $0.41\%$ for InvLT) and ties
with InvLT for the lowest NLL ($0.668$) and KDE-ECE ($0.59\%$), while InvLT
has a slightly lower AdaECE ($0.47\%$ versus $0.49\%$). DIAG comes next,
with an ECE of $0.45\%$. The input-dependent baselines PTS and MCCT lag
well behind, with ECEs of $3.92\%$ and $1.95\%$. One run of our calibrator
on ImageNet takes $880$\,s, compared with $781$\,s for InvLT and $8061$\,s
for DIAG (Appendix~\ref{app:time}); with the top-$k$ calibrator it takes
$86$\,s (Section~\ref{sec:exp-topk}).

\subsection{Exact versus soft monotonicity}
\label{sec:exp-violations}

\begin{figure}[t]
\centering
\includegraphics[width=\linewidth]{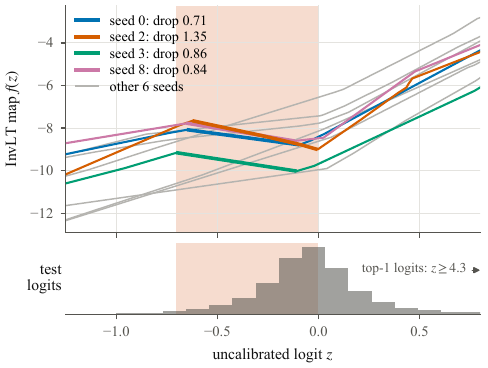}
\caption{The scalar maps $f$ learned by InvLT on ImageNet with $25000$
calibration examples, one per seed, near $z = 0$. The four colored maps
decrease on the thick segments, which lie in the shaded interval and contain
$32\%$ to $55\%$ of the test logits, depending on the seed. Of the six gray
maps, only seed $7$ decreases, by $0.002$ near $z = 0.9$. Below: distribution
of the test logits, pooled over the seeds. All top-1 logits are above $4.3$.}
\label{fig:mapping}
\end{figure}

DIAG, InvLT and our calibrator are all meant to apply a monotone map to the
logits, but only ours is monotone by construction. To check the other two,
we evaluated the scalar map $f$ that DIAG and InvLT apply to every logit on
a fine grid, for DIAG on all three datasets and for InvLT on ImageNet. With
its default quadrature, DIAG was monotone in every run, although this is
not guaranteed. InvLT was not. Figure~\ref{fig:mapping} shows its ten maps
on ImageNet with $25000$ calibration examples. Four of them decrease by
$0.7$ to $1.4$ on an interval inside $[-0.7, 0]$. This is where most
non-leading logits lie, and the interval contains $32\%$ to $55\%$ of all
test logits, so the order of many classes is reversed. Because the leading
logits lie far above it, the predictions are unaffected, and these four runs
reach ECEs between $0.29\%$ and $0.57\%$, the same range as the monotone
runs. In other words, neither the accuracy nor the calibration error shows
that the soft constraint has been violated. Our calibrator cannot learn
such a map (Proposition~\ref{prop:order}).

\subsection{Ablations}
\label{sec:exp-ablation}

\begin{table}[t]
\centering
\caption{Ablations on CIFAR-100 (ResNet-50) with $5000$ calibration
examples, averaged over $10$ seeds. ECE variants are in \%.}
\label{tab:ablation}
\setlength{\tabcolsep}{4pt}
\resizebox{\columnwidth}{!}{%
\begin{tabular}{@{}l ccccc@{}}
\toprule
Variant & NLL & ECE & AdaECE & CECE & KDE-ECE \\
\midrule
Ours (full) & 0.778 & 1.89 & 2.15 & 0.153 & 2.22 \\
\midrule
No context & 0.789 & 2.46 & 2.78 & 0.156 & 2.69 \\
No scale, $\alpha = 1$ & 0.792 & 3.24 & 3.93 & 0.161 & 3.80 \\
\bottomrule
\end{tabular}}
\end{table}

Table~\ref{tab:ablation} removes one component at a time. Without the
context, the calibrator becomes logit-wise and its CIFAR-100 ECE rises from
$1.89\%$ to $2.46\%$. Even this logit-wise variant is better calibrated than
DIAG ($3.76\%$) and InvLT ($3.43\%$). The improvement over these methods
therefore comes partly from the residual parameterization and partly from
conditioning on the input, which separates inputs that share a top logit
but differ in overconfidence (Figure~\ref{fig:gap-context}). Setting
$\alpha = 1$ hurts more, raising the ECE to $3.24\%$, so the global scale
matters more than the context.

\subsection{Calibrating only the top-\texorpdfstring{$k$}{k} logits}
\label{sec:exp-topk}

\begin{table}[t]
\centering
\caption{ImageNet (ResNet-152) with $25000$ calibration examples:
transforming all $K = 1000$ logits or only the $k = 100$ largest, averaged
over $10$ seeds. ECE variants are in \%, and Time is the wall-clock time of
one run, including evaluation. With $k = 100$, the other logits of every
method receive the clipped shift of \eqref{eq:topk}, without the scale
$\alpha$ for DIAG and InvLT.}
\label{tab:topk-main}
\setlength{\tabcolsep}{3pt}
\resizebox{\columnwidth}{!}{%
\begin{tabular}{@{}l c ccccc c@{}}
\toprule
Method & $k$ & NLL & ECE & AdaECE & CECE & KDE-ECE & Time (s) \\
\midrule
\multirow{2}{*}{DIAG} & $1000$ & 0.669 & 0.45 & 0.48 & 0.0233 & 0.60 & 8061 \\
      & $100$ & 0.670 & 0.46 & 0.48 & 0.0234 & 0.58 &  648 \\
\midrule
\multirow{2}{*}{InvLT} & $1000$ & 0.668 & 0.41 & 0.47 & 0.0233 & 0.59 &  781 \\
      & $100$ & 0.669 & 0.40 & 0.46 & 0.0232 & 0.58 &   87 \\
\midrule
\multirow{2}{*}{Ours} & $1000$ & 0.668 & 0.39 & 0.49 & 0.0233 & 0.59 &  880 \\
      & $100$ & 0.669 & 0.41 & 0.47 & 0.0233 & 0.60 &   86 \\
\bottomrule
\end{tabular}}
\end{table}

In Table~\ref{tab:imagenet}, every method transforms all $1000$ ImageNet
logits. We also ran our calibrator with the top-$k$ construction of
Section~\ref{sec:topk} and $k = 100$, and, for comparison, DIAG and InvLT
with their scalar map $f$ applied only to the $100$ largest logits. For DIAG
and InvLT, the remaining logits receive the same clipped shift as in our
calibrator, $z_j + \min\{f(\tau(z)) - \tau(z), 0\}$, detached during
training. This preserves the order between the two groups whenever $f$ is
increasing, and the test error stayed at $17.74\%$ in every run. The
results are in Table~\ref{tab:topk-main}.

The choice of $k$ has little effect on the calibration of any of the three
methods: the NLL changes by at most $0.002$ and the ECE by at most $0.02$
points. It does, however, reduce the cost considerably. With $k = 100$, all
three run $9$ to $12$ times faster, and ours and InvLT take $86$\,s and
$87$\,s. The top-$k$ construction therefore makes all three maps practical
for large label spaces, with almost no loss in calibration
(Appendix~\ref{app:topk}).

\section{Conclusion}
\label{sec:conclusion}

Existing logit-wise calibrators guarantee monotonicity only approximately and
apply the same map to every input. We proposed a post-hoc calibrator
without these limitations. It gives each input its own strictly increasing
map, selected by a permutation-invariant summary of the logits, and it
enforces monotonicity as a hard constraint, so the predictions and accuracy
of the classifier are guaranteed to stay the same. It achieves the lowest
NLL and ECE on CIFAR-10 and CIFAR-100 and matches InvLT on ImageNet. Its
main limitations are that the scale $\alpha$ is set by hand and that, within
one input, the ratio between any two slopes is bounded by $(1+L)/(1-L)$.

\section*{Impact Statement}

This paper presents work whose goal is to advance the field of Machine
Learning. There are many potential societal consequences of our work, none
which we feel must be specifically highlighted here.

\bibliography{refs}
\bibliographystyle{icml2026}

\newpage
\appendix
\onecolumn

\section{Proofs}
\label{app:proofs}

\paragraph{Proof of Lemma~\ref{lem:residual}.}
For $u \geq u'$ we have $\phi(u) - \phi(u') = (u - u') + \big(h(u) -
h(u')\big)$ and $|h(u) - h(u')| \leq L\,(u - u')$, which gives
\eqref{eq:bilip}. By the lower bound, $\phi$ is strictly increasing and
$\phi(u) \to \pm\infty$ as $u \to \pm\infty$. Since $\phi$ is continuous, it
is onto $\R$, and the lower bound applied to $\phi^{-1}$ gives the Lipschitz
bound on the inverse.
\hfill$\square$

\paragraph{Proof of Proposition~\ref{prop:topk}.}
Fix $z$, let $v = \kappa(z)$, and let $\psi(u) = u + h(u; v)$, which is
strictly increasing by Lemma~\ref{lem:residual}. Two head coordinates are
ordered correctly by Proposition~\ref{prop:order}, and two tail coordinates
because they receive the same affine map with slope $\alpha > 0$. If $z_i$ is
in the head and $z_j$ in the tail, then $z_j < \tau(z) \leq z_i$ and
\begin{equation*}
  \Phi^{k}(z)_j \;=\; \alpha\,\big(z_j + \min\{c(z), 0\}\big)
  \;\leq\; \alpha\,\big(z_j + c(z)\big)
  \;<\; \alpha\,\psi(\tau(z))
  \;\leq\; \alpha\,\psi(z_i) \;=\; \Phi^{k}(z)_i ,
\end{equation*}
so the order between head and tail is kept as well.
\hfill$\square$

\section{Evaluation metrics}
\label{app:metrics}

All metrics are computed on the test set $\{(x_n, y_n)\}_{n=1}^{N}$ after
calibration. For each test input let $p_n \in \simplex$ be the calibrated
probabilities, which equal $\sm(\Phi(z_n))$ for calibrators that act on the
logits, $\hat y_n = \arg\max_k p_{nk}$ the predicted class,
$c_n = \max_k p_{nk}$ its confidence and $a_n = \one[\hat y_n = y_n]$ its
correctness. The test error is $1 - \frac{1}{N}\sum_n a_n$. All ECE variants
are reported in percent, that is, multiplied by $100$.

\paragraph{Negative log-likelihood.}
The NLL is the mean cross-entropy of the calibrated distribution,
\begin{equation*}
  \mathrm{NLL} = -\frac{1}{N} \sum_{n=1}^{N} \log p_{n y_n},
\end{equation*}
with the natural logarithm.

\paragraph{Expected calibration error.}
The confidences are grouped into $M = 10$ bins of equal width,
$B_m = \{n : c_n \in ((m-1)/M,\ m/M]\}$ for $m = 1, \dots, M$. With the
accuracy $\mathrm{acc}(B) = \frac{1}{|B|}\sum_{n \in B} a_n$ and the mean
confidence $\mathrm{conf}(B) = \frac{1}{|B|}\sum_{n \in B} c_n$ of a bin,
\begin{equation*}
  \mathrm{ECE} = \sum_{m=1}^{M} \frac{|B_m|}{N}\,
  \big|\mathrm{acc}(B_m) - \mathrm{conf}(B_m)\big| ,
\end{equation*}
where empty bins contribute nothing \citep{guo2017calibration}.

\paragraph{Adaptive ECE.}
AdaECE \citep{nguyen2015posterior,mukhoti2020focal} uses the same formula
with $M = 10$ bins of equal mass instead of equal width. The bin edges
$e_0 \leq e_1 \leq \dots \leq e_M$ are the $m/M$ quantiles of the test
confidences, obtained by linear interpolation of the sorted values, and bin
$m$ is $\{n : c_n \in (e_{m-1}, e_m]\}$. Each bin then holds about $N/M$
inputs, whereas with equal-width bins most inputs usually fall into a few
high-confidence bins.

\paragraph{Classwise ECE.}
CECE \citep{kull2019dirichlet} measures the calibration of every class
probability, not only of the top one. For each class $k$ the probabilities
$p_{nk}$ of all test inputs are grouped into $M = 10$ bins of equal width,
$B_{km} = \{n : p_{nk} \in ((m-1)/M,\ m/M]\}$, and
\begin{equation*}
  \mathrm{CECE} = \frac{1}{K} \sum_{k=1}^{K} \sum_{m=1}^{M}
  \frac{|B_{km}|}{N}\, \bigg|\frac{1}{|B_{km}|}\sum_{n \in B_{km}} \one[y_n = k]
  - \frac{1}{|B_{km}|}\sum_{n \in B_{km}} p_{nk}\bigg| .
\end{equation*}
Since most class probabilities are close to zero, where they tend to be well
calibrated, CECE is much smaller than ECE, especially for large $K$.

\paragraph{Kernel ECE.}
KDE-ECE replaces the bins by a Gaussian kernel
$w(t, c) = \exp\!\big(-(t - c)^2 / (2h^2)\big)$ with bandwidth $h = 0.02$,
following the kernel estimate of \citet{zhang2020mixnmatch}. On $T = 200$
equally spaced points $t_1, \dots, t_T$ in $[0, 1]$ we compute the kernel
mass $D_j = \sum_n w(t_j, c_n)$ and the kernel-weighted sums
$A_j = \sum_n w(t_j, c_n)\, a_n$ and $C_j = \sum_n w(t_j, c_n)\, c_n$, and
\begin{equation*}
  \mathrm{KDE\text{-}ECE} = \frac{\sum_{j=1}^{T} |A_j - C_j|}{\sum_{j=1}^{T} D_j}
  = \sum_{j=1}^{T} \frac{D_j}{\sum_{i} D_i}
  \left|\frac{A_j}{D_j} - \frac{C_j}{D_j}\right| .
\end{equation*}
This is the average, weighted by the kernel density of the confidences, of
the gap between the smoothed accuracy $A_j / D_j$ and the smoothed confidence
$C_j / D_j$ at $t_j$. Unlike the binned estimates, it does not depend on the
choice of bin edges.

\section{Results on ImageNet}
\label{app:imagenet}

Table~\ref{tab:imagenet} compares all calibrators on ImageNet with $25000$
calibration examples; Section~\ref{sec:exp-main} discusses the results.

\begin{table}[ht]
\centering
\caption{Calibration results on ImageNet (ResNet-152) with $25000$
calibration examples, mean and standard deviation over $10$ seeds. All ECE
variants are in \%. Best mean among the calibrators in \textbf{bold}, second
best \underline{underlined}; the uncalibrated model is excluded from the
ranking.}
\label{tab:imagenet}
\begin{tabular}{l ccccc}
\toprule
Method & NLL & ECE & AdaECE & CECE & KDE-ECE \\
\midrule
Uncalibrated & 0.869$_{\pm 0.005}$ & 12.80$_{\pm 0.15}$ & 12.80$_{\pm 0.15}$ & 0.0393$_{\pm 0.0002}$ & 12.82$_{\pm 0.16}$ \\
\midrule
TS & 0.741$_{\pm 0.005}$ & 3.97$_{\pm 0.18}$ & 4.20$_{\pm 0.20}$ & 0.0237$_{\pm 0.0002}$ & 4.16$_{\pm 0.19}$ \\
VS & 0.833$_{\pm 0.019}$ & 4.84$_{\pm 0.23}$ & 4.84$_{\pm 0.22}$ & 0.0238$_{\pm 0.0003}$ & 5.18$_{\pm 0.27}$ \\
MS & 0.735$_{\pm 0.007}$ & 4.42$_{\pm 0.23}$ & 4.42$_{\pm 0.23}$ & 0.0237$_{\pm 0.0003}$ & 4.73$_{\pm 0.26}$ \\
HB & 2.390$_{\pm 0.054}$ & 4.50$_{\pm 0.35}$ & 6.73$_{\pm 0.23}$ & 0.0243$_{\pm 0.0005}$ & 5.91$_{\pm 0.14}$ \\
IR & 1.462$_{\pm 0.035}$ & 3.32$_{\pm 0.18}$ & 3.32$_{\pm 0.18}$ & \underline{0.0231}$_{\pm 0.0003}$ & 3.67$_{\pm 0.21}$ \\
Spline & 0.711$_{\pm 0.006}$ & 1.78$_{\pm 0.10}$ & 1.68$_{\pm 0.13}$ & \textbf{0.0229}$_{\pm 0.0002}$ & 1.95$_{\pm 0.12}$ \\
Dirichlet & 0.760$_{\pm 0.009}$ & 5.61$_{\pm 0.22}$ & 5.60$_{\pm 0.22}$ & 0.0243$_{\pm 0.0003}$ & 6.15$_{\pm 0.25}$ \\
ETS & 0.741$_{\pm 0.005}$ & 3.97$_{\pm 0.18}$ & 4.20$_{\pm 0.20}$ & 0.0237$_{\pm 0.0002}$ & 4.16$_{\pm 0.19}$ \\
PTS & 0.736$_{\pm 0.009}$ & 3.92$_{\pm 0.19}$ & 4.10$_{\pm 0.27}$ & 0.0236$_{\pm 0.0003}$ & 4.18$_{\pm 0.18}$ \\
DIAG & 0.669$_{\pm 0.005}$ & 0.45$_{\pm 0.15}$ & \underline{0.48}$_{\pm 0.14}$ & 0.0233$_{\pm 0.0001}$ & 0.60$_{\pm 0.13}$ \\
MCCT & 0.681$_{\pm 0.005}$ & 1.95$_{\pm 0.21}$ & 1.96$_{\pm 0.21}$ & 0.0236$_{\pm 0.0002}$ & 1.94$_{\pm 0.21}$ \\
InvLT & \textbf{0.668}$_{\pm 0.005}$ & \underline{0.41}$_{\pm 0.12}$ & \textbf{0.47}$_{\pm 0.09}$ & 0.0233$_{\pm 0.0001}$ & \textbf{0.59}$_{\pm 0.09}$ \\
\midrule
Ours & \textbf{0.668}$_{\pm 0.005}$ & \textbf{0.39}$_{\pm 0.07}$ & 0.49$_{\pm 0.11}$ & 0.0233$_{\pm 0.0001}$ & \textbf{0.59}$_{\pm 0.08}$ \\
\bottomrule
\end{tabular}
\end{table}

\section{Calibrating only the top-\texorpdfstring{$k$}{k} logits}
\label{app:topk}

\begin{table}[ht]
\centering
\caption{Calibrating only the top-$k$ logits on ImageNet (ResNet-152) with
$25000$ calibration examples, averaged over $10$ seeds. ``Logits'' is the
number of logits per example at which the residual network is evaluated.
``Fit'' is the time to fit the calibrator and ``Run'' the time of a whole
run, including loading the logits and evaluation, both on one CPU core (AMD
EPYC 9374F), with the two settings timed in the same batch of runs. ECE variants are in \%. Temperature scaling is shown for
reference.}
\label{tab:topk}
\setlength{\tabcolsep}{4.5pt}
\begin{tabular}{l ccc ccccc}
\toprule
Method & Logits & Fit (s) & Run (s) & NLL & ECE & AdaECE & CECE & KDE-ECE \\
\midrule
TS & -- & -- & -- & 0.741 & 3.97 & 4.20 & 0.0237 & 4.16 \\
\midrule
Ours, $k = K = 1000$ & 1000 & 725.3 & 736.5 & 0.668 & 0.39 & 0.49 & 0.0233 & 0.59 \\
Ours, $k = 100$ & 100 & 75.7 & 85.6 & 0.669 & 0.41 & 0.47 & 0.0233 & 0.60 \\
\bottomrule
\end{tabular}
\end{table}

Table~\ref{tab:topk} compares the top-$k$ calibrator of
Section~\ref{sec:topk} with $k = 100$ to the full calibrator
($k = K = 1000$) on ImageNet with $25000$ calibration examples, as in
Table~\ref{tab:imagenet}; all other settings follow
Section~\ref{sec:exp-setup}. With $k = 100$, the residual network is
evaluated at a tenth of the logits, and fitting takes $75.7$\,s instead of
$725.3$\,s, a $9.6$-fold speed-up that is close to the ratio $K/k = 10$. A
whole run is $8.6$ times faster, because loading the logits and computing
the metrics do not depend on $k$. The NLL increases by $0.0005$ and the ECE
by $0.02$ percentage points, while the AdaECE and CECE stay the same or
improve slightly. Most of the benefit of the input-dependent map therefore
comes from the leading logits, and the tail can be handled by the clipped
common shift at almost no cost.

\section{Choice of context}
\label{app:context}

Table~\ref{tab:context} varies the context of our calibrator and, for
comparison, of PTS, which predicts one temperature per input from the same
kind of summary. The context consists of the $m$ largest softmax
probabilities or the $m$ largest logits, sorted in decreasing order, with
$m \in \{1, 2, 3, 5, 10\}$; all other settings are as in
Section~\ref{sec:exp-setup}.

Which kind of context works better depends on the dataset. Logits are better
on CIFAR, most clearly on CIFAR-100, where three logits give an ECE of
$1.14\%$ compared with $1.89\%$ for our default of two probabilities.
Probabilities are better on ImageNet, where two probabilities give $0.41\%$
compared with $0.53\%$ for the best logit context, and logit contexts with
$m \geq 5$ degrade. On CIFAR, probabilities are also less sensitive to $m$.
On ImageNet, one of the ten runs with five probabilities failed, reaching an
ECE of $15\%$, which pulls the mean for that setting up to $1.86\%$. Ten
logits are clearly worse on all three datasets. In the main experiments we
use two probabilities on every dataset instead of tuning the context per
dataset, because probabilities are bounded, unaffected by a common shift of
the logits, and best on the largest dataset. On CIFAR-100, every context in
the table still gives better calibration than all baselines in
Table~\ref{tab:main}.

PTS, on the other hand, is almost insensitive to the context: its ECE stays
between $4.15\%$ and $4.41\%$ on CIFAR-100 and between $3.85\%$ and $3.97\%$
on ImageNet for every choice. PTS receives the same information as our
calibrator but applies a linear map within each input. The contrast between
the two halves of the table therefore suggests that a per-input temperature
is limited mainly by being linear within an input, rather than by a lack of
information about the input.

\begin{table}[ht]
\centering
\caption{ECE (\%) of our calibrator and of PTS for different contexts, with
$5000$ calibration examples on CIFAR and $25000$ on ImageNet, averaged over
$10$ seeds. The context is the
$m$ largest entries of the given type, sorted in decreasing order. Our
default is \underline{underlined}. On ImageNet these runs use the top-$k$
calibrator with $k = 100$.
``Probs'' are softmax probabilities.}
\label{tab:context}
\setlength{\tabcolsep}{4.5pt}
\begin{tabular}{l c ccc ccc}
\toprule
& & \multicolumn{3}{c}{Ours} & \multicolumn{3}{c}{PTS} \\
\cmidrule(lr){3-5}\cmidrule(lr){6-8}
Context & $m$ & CIFAR-10 & CIFAR-100 & ImageNet & CIFAR-10 & CIFAR-100 & ImageNet \\
\midrule
Probs & 1 & 0.53 & 1.95 & 0.52 & 0.96 & 4.19 & 3.96 \\
 & 2 & \underline{0.60} & \underline{1.89} & \underline{0.41} & 0.96 & 4.20 & 3.91 \\
 & 3 & 0.59 & 1.84 & 0.44 & 0.95 & 4.16 & 3.94 \\
 & 5 & 0.63 & 1.68 & 1.86 & 0.97 & 4.15 & 3.96 \\
 & 10 & 0.53 & 1.90 & 0.46 & 0.95 & 4.15 & 3.91 \\
\midrule
Logits & 1 & 0.50 & 1.43 & 0.53 & 1.04 & 4.41 & 3.97 \\
 & 2 & 0.58 & 1.23 & 0.59 & 0.85 & 4.29 & 3.90 \\
 & 3 & 0.53 & 1.14 & 0.56 & 0.85 & 4.22 & 3.85 \\
 & 5 & 0.59 & 1.13 & 0.95 & 0.85 & 4.29 & 3.95 \\
 & 10 & 0.93 & 2.74 & 1.24 & 0.87 & 4.32 & 3.92 \\
\bottomrule
\end{tabular}
\end{table}

\section{Running time}
\label{app:time}

Table~\ref{tab:time} reports the wall-clock time of one run of each
order-preserving calibrator, with $5000$ calibration examples on CIFAR and
$25000$ on ImageNet. A run consists of fitting the calibrator and evaluating
it on the test set. All runs use the same code base and settings as
Tables~\ref{tab:main} and~\ref{tab:imagenet}. The cost of DIAG is dominated
by the numerical integration, which is repeated at every logit of every
example in each training step. On ImageNet, our calibrator, which here
transforms all $1000$ logits, takes slightly longer than InvLT and about a
ninth of the time of DIAG. With the top-$k$ calibrator and $k = 100$ it
takes $86$\,s (Table~\ref{tab:topk-main}).

\begin{table}[ht]
\centering
\caption{Wall-clock time of one run (fitting and evaluation) in seconds, with
$5000$ calibration examples on CIFAR and $25000$ on ImageNet, averaged over
$10$ seeds.}
\label{tab:time}
\begin{tabular}{l ccc}
\toprule
Method & CIFAR-10 & CIFAR-100 & ImageNet \\
\midrule
TS     & 1.7 & 2.3  & 8.3 \\
ETS    & 1.0 & 1.7  & 5.2 \\
PTS    & 1.4 & 2.7  & 52.6 \\
DIAG   & 8.1 & 90.9 & 8061.3 \\
MCCT & 1.3 & 10.2 & 713.6 \\
InvLT  & 1.9 & 15.4 & 781.1 \\
\midrule
Ours   & 1.9 & 16.7 & 880.1 \\
\bottomrule
\end{tabular}
\end{table}

\section{Reliability diagrams}
\label{app:reliability}

\begin{figure}[ht]
\centering
\includegraphics{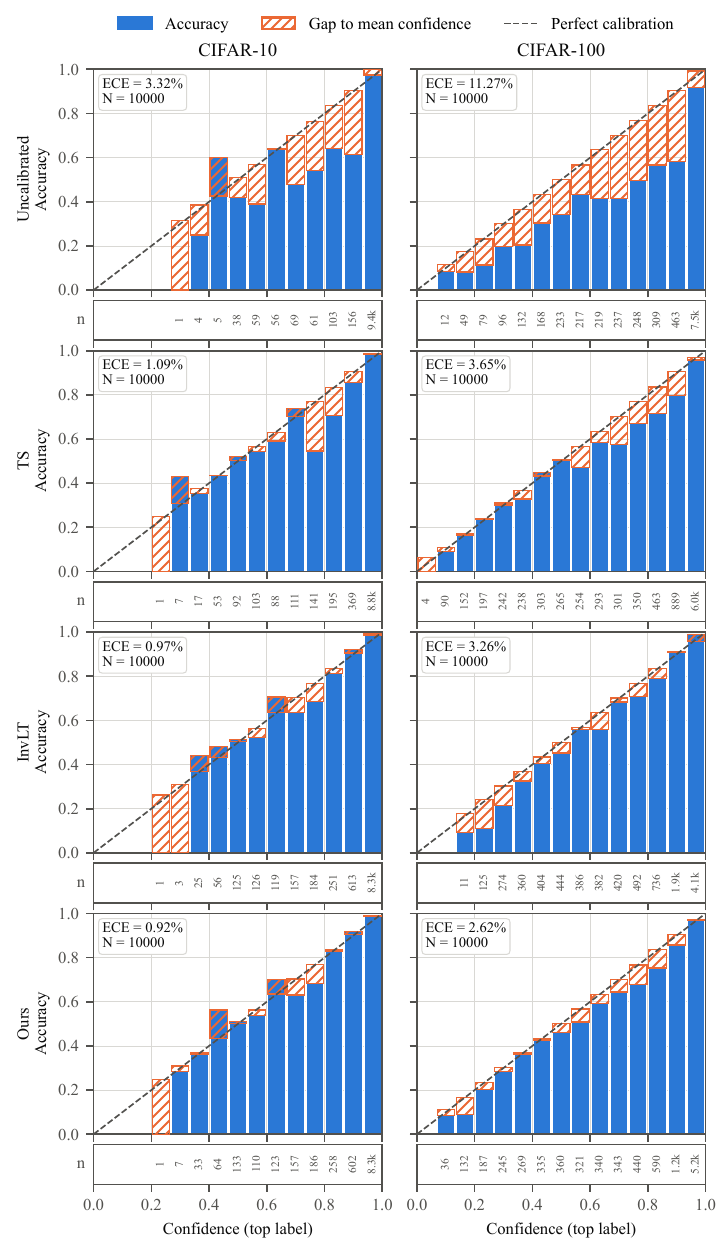}
\caption{Reliability diagrams on the CIFAR-10 and CIFAR-100 test sets ($15$
equal-width bins).
Bars show the accuracy in each bin, hatched boxes the gap to the mean
confidence, and the panels below each diagram the number of test inputs per
bin.}
\label{fig:reliability}
\end{figure}

Figure~\ref{fig:reliability} shows reliability diagrams of the top-label
confidence for the uncalibrated classifier, temperature scaling, InvLT and our
calibrator on CIFAR-10 and CIFAR-100, with $5000$ calibration examples and
seed $0$. Without calibration, both models are overconfident. Temperature
scaling removes most of the gap but leaves the CIFAR-100 model overconfident
in the middle of the confidence range, where a single temperature cannot
suit both confident and uncertain inputs. InvLT and our calibrator stay
close to the diagonal there, with smaller gaps for ours ($15$-bin ECE of
$2.62\%$ versus $3.26\%$).

\end{document}